\documentclass[a4paper]{article}

\usepackage{INTERSPEECH2021}

\newcolumntype{C}{>{\centering\arraybackslash}X}
\newcolumntype{L}{>{\raggedright\arraybackslash}X}
\newcolumntype{R}{>{\raggedleft\arraybackslash}X}

\title{Attention-Based Keyword Localisation in Speech using Visual Grounding}
\name{Kayode Olaleye \qquad Herman Kamper}
\address{E\&E Engineering, Stellenbosch University} 
\email{kaykola.olaleye@gmail.com, kamperh@sun.ac.za}

\usepackage{microtype}
\usepackage[bottom]{footmisc}

\usepackage[prependcaption,textsize=scriptsize]{todonotes}
\definecolor{mycolor}{HTML}{FF6600}

\begin{document}

\maketitle
\begin{abstract}
Visually grounded speech models learn from images paired with spoken captions. By tagging images with soft text labels using a trained visual classifier with a fixed vocabulary, previous work has shown that it is possible to train a model that can \textit{detect} whether a particular text keyword occurs in speech utterances or not. Here we investigate whether visually grounded speech models can also do keyword \textit{localisation}: predicting where, within an utterance, a given textual keyword occurs without any explicit text-based or alignment supervision. We specifically consider whether incorporating attention into a convolutional model is beneficial for localisation. Although absolute localisation performance with visually supervised models is still modest (compared to using unordered bag-of-word text labels for supervision), we show that attention provides a large gain in performance over previous visually grounded models. As in many other speech-image studies, we find that many of the incorrect localisations are due to semantic confusions, e.g.\ locating the word `backstroke' for the query keyword `swimming'.
\end{abstract}


\begin{figure*}[!t]
  \centering
  \includegraphics[width=0.95\linewidth]{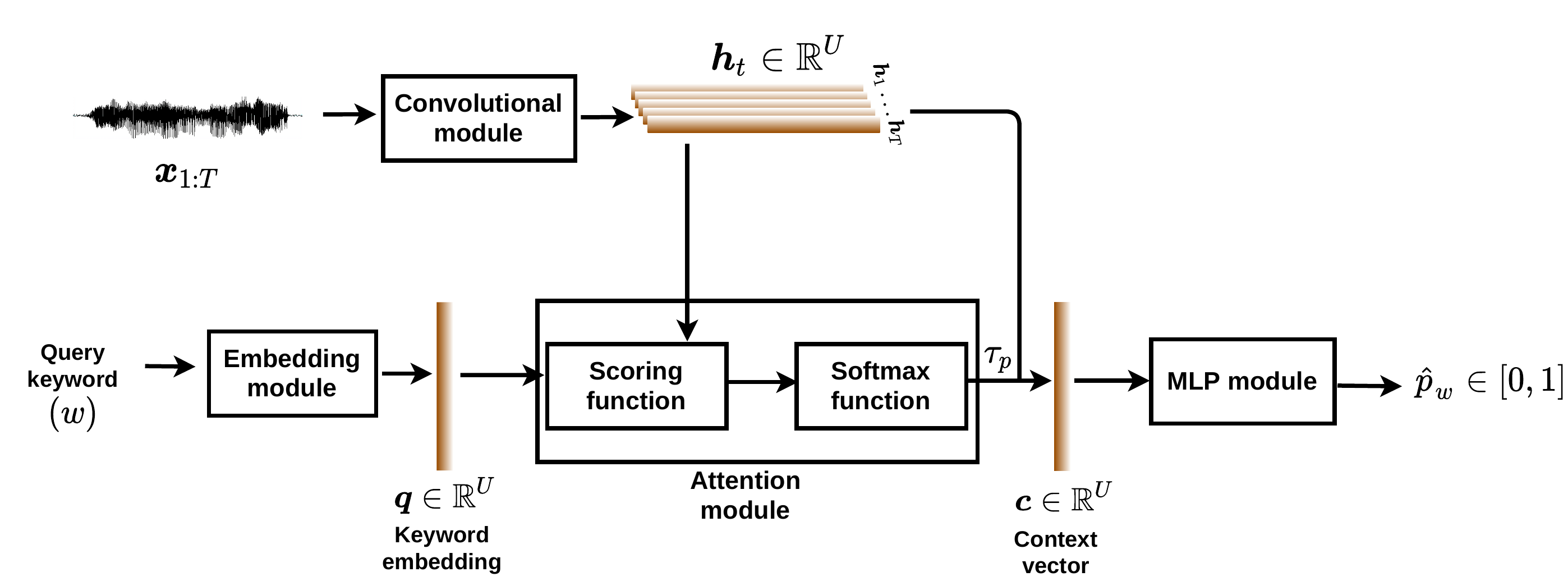}
    \vspace*{-5pt}
  \caption{The structure of our attention-based CNN for keyword localisation in speech.}
  \label{fig:speech_production}
    \vspace*{-1pt}
\end{figure*}

\section{Introduction}


Several studies have considered how speech processing systems can be developed in the absence of conventional transcriptions~\cite{driesen2010, synnaeve2014b, duong2016, palaz2016, settle2017, weiss2017}.
This could enable speech technology in low-resource settings where full transcriptions are not available.
One form of weak supervision is using images paired with spoken captions~\cite{harwath2015, harwath2016, kamper2017, harwath2017a, harwath2018a, kamper2018, harwath2018b, eloff2019, harwath2019a, harwath2019b}.
Compared to using labelled data, this form of visual supervision is closer to the types of signals that
infants would have access to while learning their first language~\cite{bomba1983, pinker1994, eimas+quinn94, roy2003, boves2007, chrupala2016}, and to how one would teach new words to robots using spoken language~\cite{meng2013, nortje2020}.
It is also conceivable that this type of visual supervision could be easier to obtain when developing systems for low-resource languages~\cite{de1998}, e.g.\ when it is not possible to collect textual labels for a language without a written form.

In this visually supervised setting, both the utterance and the paired image are unlabelled.
Some form of self-supervision 
is therefore required to train a model.
One approach is to train a model that projects images and speech into a joint embedding space~\cite{harwath2016,chrupala2017}.
Another approach is to use an external image tagger with a fixed vocabulary to obtain soft text labels, which are then used as targets for a speech network that maps speech to keyword labels~\cite{kamper2019b, pasad2019}.
The advantage of this latter approach is that it can be used to \textit{detect} whether a textual keyword is present in a new input utterance (as long as the keyword is in the vocabulary).
Without seeing any transcriptions, previous work has shown that such visually grounded models can do keyword detection at a high precision~\cite{kamper2017}.
However, current approaches does not indicate \textit{where} in an utterance a detected keyword occurs.
In this paper we investigate whether visually grounded speech models can be used for this task of \textit{keyword localisation}: predicting where in an utterance a given textual keyword occurs.
The model needs to do so without any explicit text labels or word position supervision.

In our own initial efforts toward this goal~\cite{olaleye2020}, we tried to adapt a saliency-based method often used in the vision community for determining which parts of an input image contributes most to a particular output prediction~\cite{selvaraju2017}.
We found, however, that this approach was poorly matched to the particular multi-label loss used for the visually grounded model.
More specifically, this approach gave worse localisation performance than a visually grounded version of what we call the PSC model~\cite{palaz2016} (based on the author names).
The original PSC 
uses a convolutional neural network (CNN) architecture that is tailored to jointly locate and classify words using a bag-of-words (BoW) labelling, i.e.\ the model is trained with labels indicating the presence or absence of a word but not the location, order, or number of occurrences.
We adapted this model to use visual supervision. 
Although the visually grounded PSC model gave better localisation scores relative to the saliency-based approach, absolute performance was still low.

In this paper we specifically consider whether incorporating \textit{attention} into a visually grounded CNN is beneficial for localisation.
Attention has become a standard mechanism to allow a neural network to automatically weigh different parts of its input when producing some (intermediate) output~\cite{luong2015, chan2016}.
In our architecture, a given textual keyword is mapped to a learned query embedding, which is then used to weigh the relevance of intermediate convolutional filters.
The model is trained using a multi-label loss using the soft tags from an external visual classifier.
This architecture is heavily inspired by~\cite{tamer2020}, which considered supervised keyword localisation in sign language; instead of using text labels, here we use visual grounding.

Without using any explicit text or alignment supervision, we show that an attention-based model 
outperforms the visually grounded PSC by roughly 17\% absolute in localisation $F1$.
However, the absolute performance is still modest.
We subsequently do an error analysis to determine the types of localisation mistakes that the model makes.
As in previous studies on visually grounded speech models~\cite{kamper2017, kamper2019b}, we find that in many cases the model is penalised for locating a word which is not an exact match to the query but is semantically related.

\section{CNNs for keyword localisation in speech}
\label{sect:models}

\subsection{Attention-based CNN for keyword localisation}

\label{sect:attention_based_model}
Our attention-based CNN is heavily inspired by \cite{tamer2020}, which proposed an attention-based graph-convolutional network over skeleton joints to search for keywords in sign language sentences.
In our case, we incorporate 
attention 
in a CNN model for
localising keywords in speech, and train the model with visual supervision instead of explicit labels for whether a keyword is present in an utterance or not.
Our architecture, shown in Figure~\ref{fig:speech_production},
consists of four modules: a \textit{convolutional module} to extract features from speech, an \textit{embedding module} to represent 
a query keyword, an 
\textit{attention module} to compress the convolutional features into a context vector 
containing features most relevant to reaching a decision about the query keyword, and a \textit{multi-layer perceptron} (MLP) module that takes the output
of the attention module and produces a probability 
about the presence or absence of the query keyword 
in the speech.

The convolutional module consists of a succession of one-dimensional convolutional layers that produces the intermediate features that are weighed using the query embedding vector.
This module takes as input a sequence of $T$ Mel-frequency cepstral coefficient 
vectors $\boldsymbol{X} = (\boldsymbol{x}_1, \ldots, \boldsymbol{x}_T)$ and outputs a sequence of convolutional features $\boldsymbol{h}_{1:T}$, each of dimension $U$.

The embedding module takes a text keyword $w$ as input and maps it to a $U$-dimensional query vector $\boldsymbol{q}$ using an index-based lookup table. $U$ is selected to match the dimension of the sequence of convolutional features $\boldsymbol{h}_{1:T}$.
The type of keyword needs to be in the vocabulary of the system, and the embedding matrix is trained jointly with the rest of the model.

The attention module \cite{graves2014,luong2015} starts by computing a score $e_t$ for each time step $t$ by using a dot product to measure the similarity between each time step along the convolutional features $\vec{h}_t$ and the query 
embedding: $e_t = \boldsymbol{q} \cdot \boldsymbol{h}_t$.
The similarity score $e_t$ is then converted into an attention weight $\alpha_t$ over time steps using a softmax function: $\alpha_t = \frac{\exp(e_t)}{\sum_{{t' = 1}}^T\exp(e_{t'})}$. The attention weight is used to compute the context vector $\boldsymbol{c}$ by weighing the features $\boldsymbol{h}_t$ at every time step: $\boldsymbol{c} = \sum_{t = 1}^T \alpha_t \boldsymbol{h}_t$. 

Finally, the MLP module consumes 
$\boldsymbol{c}$, passes it through a number of fully-connected layers, and terminates with a layer with a sigmoid activation producing a single probability output indicating the presence or absence of the keyword, i.e.\ $\hat{p}_w = \sigma(\text{MLP}(\vec{c}))$ and we interpret $\hat{p}_w \in [0, 1]$ as $P(w\vert \vec{X})$.

How is the model trained?
Let's say we know at training time whether a particular keyword $w$ occurs in an output utterance $\boldsymbol{X}$, i.e.\ we have an indicator variable $y_{\text{bow}, w} \in \{0, 1\}$. (But we still don't know where $w$ occurs).
In this case we could train the model with the binary cross-entropy loss:
\begin{equation*} J(\hat{{p}}_w, {y}_{\text{bow}, w}) = y_{\text{bow},w} \log \hat{p}_w + (1 - y_{\text{bow},w}) \log (1 - \textrm{log} \hat{p}_w)
\label{binary_log_loss_eqn}
\end{equation*}
where $\hat{p}_w$ is the output of the model for a single training example $\boldsymbol{X}$ and we look up the $w^{\text{th}}$ embedding vector $\boldsymbol{q}$ for the attention model.
When training with paired images and spoken captions, we don't actually know whether $w$ occurs in the training utterance $\boldsymbol{X}$.
So 
we use the above loss but replace ${y}_{\text{bow}, w}$ with ${y}_{\text{vis}, w} \in [0, 1]$, 
the soft probability obtained for keyword $w$ by passing the paired image through an external visual tagger.

How can the model detect whether a keyword is present or absent in a new test utterance? At test time, a threshold $\theta$ can be applied to the output of the model $\hat{p}_w$ to determine the presence or absence of the query keyword $w$. 
How can the model locate where a detected keyword occurs?
At test time, we select the position of the highest attention weight $\alpha_t$ produced by the model as the predicted location $\tau_p$ of a detected keyword $w$.

\subsection{Model variants}
\label{subsec:model_variants}
For the convolutional module, we consider two different structures. The first, based on~\cite{palaz2016}, consist of six convolutional layers without any intermediate max-pooling.
We refer to the resulting model that uses this convolutional module together with the query and attention modules as \textit{CNN-Attend}.
The second structure, based on~\cite{kamper2017}, uses three convolutional layers with intermediate max-pooling. We refer to the resulting model as \textit{CNN-PoolAttend}.

\subsection{Baseline models}
\label{subsec:base_model}
The two architectures for the convolutional module above are actually based on two non-attention CNN models from previous work, which also serves as our baselines.
We refer to the first as PSC~\cite{palaz2016}, which is the non-attention parallel to the CNN-Attend model. PSC was originally designed to jointly locate and classify words using only BoW supervision.
It operates on $\boldsymbol{X}$ using a stack of convolutional layers and an aggregation function. It performs keyword localisation using the output of its last convolutional layer $\boldsymbol{h}_{1:T}$. 
Here, each $\vec{h}_t$ 
is a $W$-dimensional vector, with {$h_{t, w} = \alpha_{t, w}$} giving the score for word $w$ at time $t$. $W$ is the number of words in the vocabulary.
A single utterance-level detection score is obtained by feeding the frame-level scores into the aggregation function: $g_w(\boldsymbol{X}) = \frac{1}{r} \log \left[ \frac{1}{T} \sum^T_{t = 1} \exp \left\{r h_{t, w}(\boldsymbol{X})\right\} \right]$.
Finally, a sigmoid function converts the speech-level scores into a probability $\hat{\boldsymbol{y}} = \sigma ( \boldsymbol{g}(\boldsymbol{X}))$, with each $\hat{y}_w$ giving the probability that word $w$ occurs in $\vec{X}$.
The model is trained with the cross-entropy loss:
\begin{align}
&J(\hat{\boldsymbol{{y}}}, \boldsymbol{y}_\text{bow}) = \nonumber \\ 
&\ \ -\sum_{w=1}^W \left\{ y_{\text{bow},w} \log \hat{y}_w + (1 - y_{\text{bow},w}) \log (1 - \textrm{log} \hat{y}_w) \right\}
\label{binary_log_loss_eqn}
\end{align}
where $\boldsymbol{y}_{\text{bow}} \in \{0, 1 \}^W$ indicates that BoW supervision is used. For visual supervision, it is replaced by $\boldsymbol{y}_{\text{vis}} \in [0, 1]^W$, which is the vector of soft probabilities for the $W$ words in the output of the multi-label visual tagger.

Our second non-attention baseline is \textit{CNN-Pool}, the parallel to \textit{CNN-PoolAttend}. It
uses a number of convolutional layers with intermediate max-pooling, max-pooling over time over its last convolutional layer, and a number of fully connected layers before terminating in a sigmoid output.
The output is again denoted as $\hat{\boldsymbol{y}}$ and the same loss as in~\eqref{binary_log_loss_eqn}
is used.
While PSC has a localisation mechanism built into the architecture, this is not the case for CNN-Pool. As in~\cite{olaleye2020}, we therefore use the GradCAM~\cite{selvaraju2017} saliency-based localisation method to do localisation. In short, this works by first determining how important each filter in the output of the last one-dimensional convolutional layer $\boldsymbol{h}_{1:T}$
is to the word $w$, given by $\gamma_{k, w} = \frac{1}{T} \sum^T_{t = 1} \frac{\partial \hat{y}_w}{\partial h_{t, k}}$. Each filter is then weighed by its importance, giving 
the localisation scores {$\alpha_{t, w} = \textrm{ReLU} \left[ \sum_{k = 1}^K \gamma_{k, w} h_{t, k} \right]$}.

\section{Experimental setup}

\subsection{Data}

We perform experiments on the Flickr8k Audio Caption Corpus of~\cite{harwath2015}, consisting  of five English audio captions for each of the roughly $8$k images. A split of $30$k, $5$k and $5$k 
utterances 
are used for training, development and testing, respectively.

The visually grounded models are trained using soft text labels obtained by passing each training image through the multi-label visual tagger of~\cite{kamper2019b}, which is based on VGG-16~\cite{simonyan2014} and is trained on images disjoint from the data used here~\cite{imagenet2009,krizhevsky2012,girshick2014}.
This tagger has an output of 1000 image classes, but here we use a system vocabulary corresponding to $W = 67$ unique keyword types.
This set of keywords is the same set used in~\cite{kamper2019b, kamper2019a}, and includes words such as `children', `young', `swimming', `snowy' and `riding'.
The procedure 
used to select these keywords are detailed in~\cite{kamper2019b}; it includes a human reviewer agreement step, which reduced the original set from 70 to 67 words.

\begin{figure}[!b]
    \centering
    \includegraphics[width=0.99\linewidth]{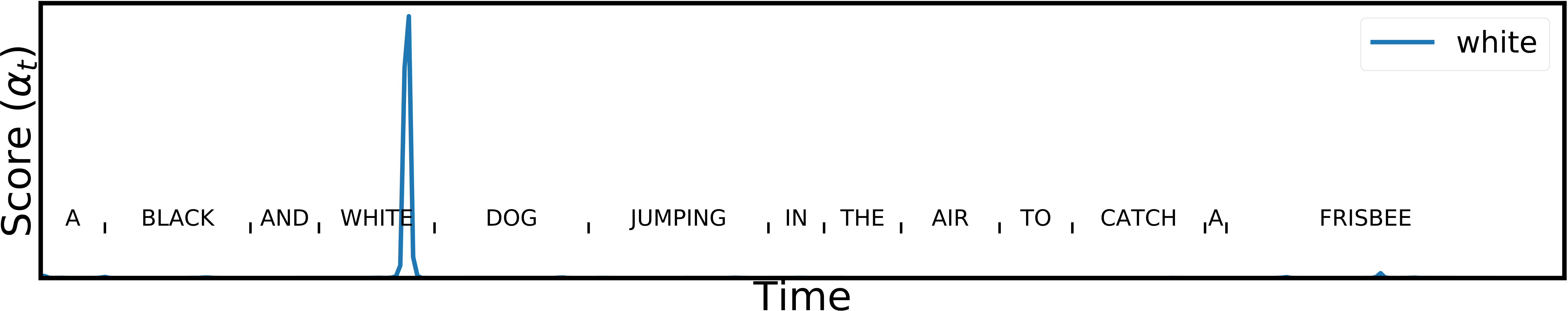} \\[-3pt]
    {\footnotesize (a)} \\[5pt]
    
    \includegraphics[width=0.99\linewidth]{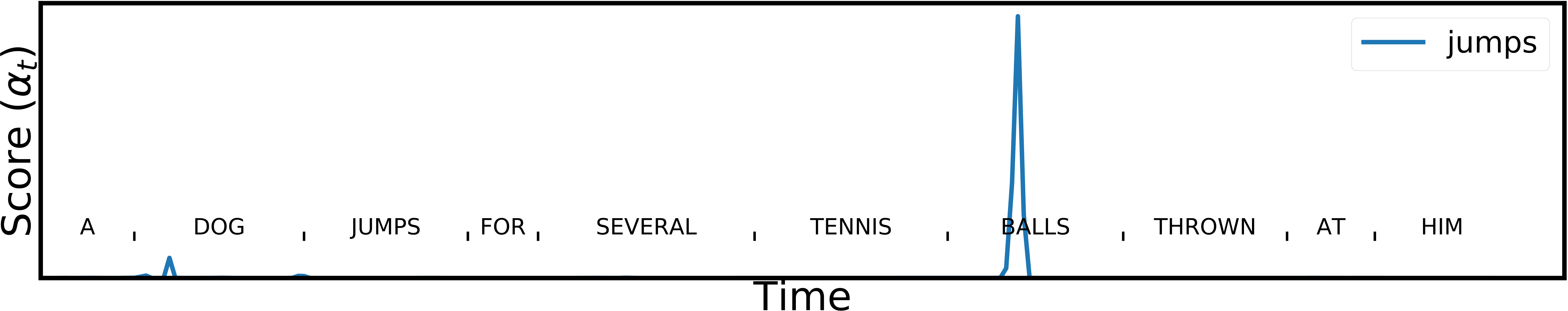} \\[-3pt]
    {\footnotesize (b)}
    
    \includegraphics[width=0.99\linewidth]{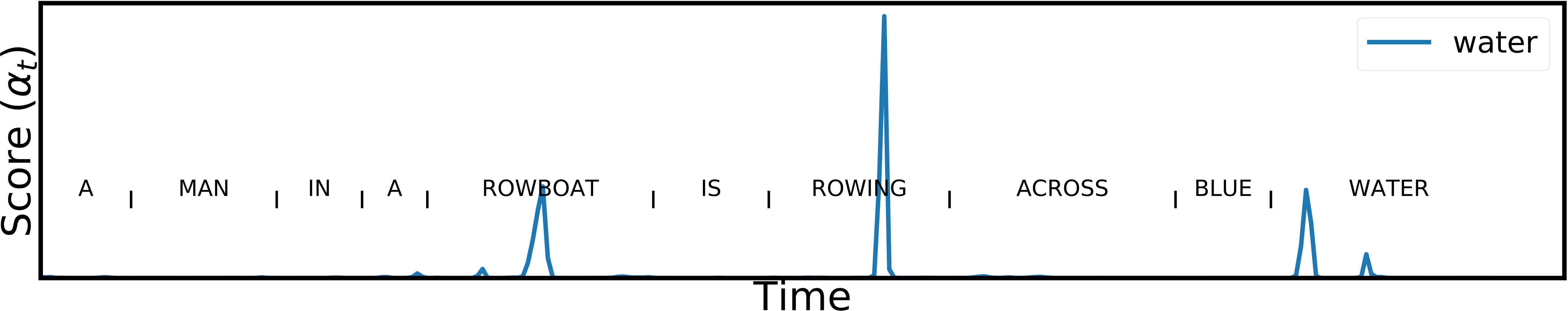} \\[-3pt]
    {\footnotesize (c)}
    
    \includegraphics[width=0.99\linewidth]{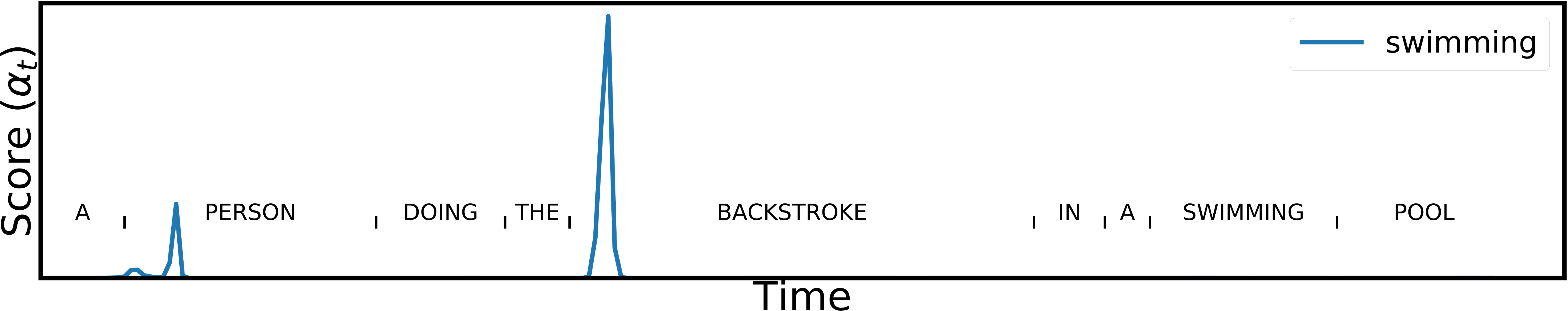} \\[-3pt]
    {\footnotesize (d)} 
    
   	\vspace*{-5pt}
    \caption{Examples of localisation with the visually supervised CNN-Attend model. The query keyword 
        is shown on the right. 
    }
    \label{fig:qualitative_eval}
\end{figure}

\begin{table*}[t]
  \captionof{table}{
      Keyword localisation scores (\%), where the task is to predict where an utterance a given keyword occurs.
} 
   	\vspace*{-5pt}
\label{tbl:localisation_evaluation}
	\centering
	\renewcommand{\arraystretch}{1.1}
   \begin{tabularx}{0.8\textwidth}{@{}lCCCCCC@{}} 
    \toprule
      & \multicolumn{3}{c}{Bag-of-words supervision} & \multicolumn{3}{c}{Visual supervision}\\
       \cmidrule(lr){2-4}\cmidrule(l){5-7}
      Model & $P$ & $R$ & $F1$ & $P$ & $R$ & $F1$\\
      \midrule
      \multicolumn{3}{@{}l}{\textit{Attention-based models:}}\\
      \textbf{CNN-Attend} & $58.1$ & $68.1$ & $62.7$ & $31.2$ & $15.0$ & $20.3$\\
      \textbf{CNN-PoolAttend} & $30.8$ & $47.2$ & $37.3$ & $10.0$ & $8.0$ & $8.9$\\[2pt]
      \multicolumn{3}{@{}l}{\textit{Baseline models:}}\\
      \textbf{PSC} & $78.4$ & $70.4$ & $74.2$ & $7.6$ & $2.2$ & $3.4$\\
      \textbf{CNN-Pool} & $10.3$ & $32.1$ & $15.7$ & $7.1$ & $10.8$ & $8.6$\\
      \bottomrule
    \end{tabularx}
\end{table*}

\begin{table*}[t]
  \captionof{table}{
  Keyword detection scores (\%), where the task is to detect whether a given keyword occurs in an utterance (regardless~of~location).
  } 
\label{tbl:detection_evaluation}
   	\vspace*{-5pt}
	\centering
	\renewcommand{\arraystretch}{1.1}
   \begin{tabularx}{0.8\textwidth}{@{}lCCCCCC@{}} 
    \toprule
      & \multicolumn{3}{c}{Bag-of-words supervision} & \multicolumn{3}{c}{Visual supervision}\\
       \cmidrule(lr){2-4}\cmidrule(l){5-7}
      Model & $P$ & $R$ & $F1$ & $P$ & $R$ & $F1$\\
      \midrule
      \multicolumn{3}{@{}l}{\textit{Attention-based models:}}\\
      \textbf{CNN-Attend} & $82.8$ & $79.2$ & $80.9$ & $30.6$ & $27.2$ & $28.8$\\
      \textbf{CNN-PoolAttend} & $74.0$ & $70.6$ & $72.3$ & $18.6$ & $23.0$ & $20.5$\\[2pt]
      \multicolumn{3}{@{}l}{\textit{Baseline models:}}\\
      \textbf{PSC} & $86.9$ & $70.4$ & $77.8$ & $20.8$ & $7.0$ & $10.4$\\
      \textbf{CNN-Pool} & $81.5$ & $73.0$ & $77.0$ & $33.1$ & $26.2$ & $29.2$\\
      \bottomrule
    \end{tabularx}
\end{table*}

\subsection{Model architectures} 

The $W = 67$ soft probabilities from the tagger are 
used as the targets $\vec{y}_{\text{vis}}$  
for the visually grounded models.
As an upper limit for how well we can expect these visual 
models to perform,
we train 
models with similar structures but using 
BoW labels
$\vec{y}_\text{bow}$ 
obtained using transcriptions for the same set of keywords. 

We first look at the baseline models (Section~\ref{subsec:base_model}).
The PSC model consists of six one-dimensional convolutional layers with ReLU activations. The first has $96$ filters with a kernel width of $9$ frames. The next four has a width of $11$ units with $96$ filters. The last convolutional layer, with 
{$1000$} filters and a width of $11$ units, is fed into the aggregation function with a final sigmoid activation. The CNN-Pool model consists of three one-dimensional convolutional layers with ReLU activations. Intermediate max-pooling over 3 units are applied in the first two layers. The first convolutional layer has $64$ filters with a width of $9$ frames. The second layer has $256$ filters with a width of $11$ units, and the last layer has $1024$ filters with a width of $11$. Global max-pooling is applied followed by two fully connected layers each with $4096$ and $1000$ filters respectively and terminates in sigmoid activation to obtain the final output for the $W = 67$ words. 

Next we turn to our attention-based models (Section~\ref{subsec:model_variants}).
The convolutional module of CNN-Attend has the same structure as the PSC model above while the convolutional module of CNN-PoolAttend has the same structure as CNN-Pool. The embedding module embeds input query keywords into $U=1000$ dimensional embeddings in CNN-Attend and $U=1024$ dimensional embeddings in CNN-PoolAttend.
For both models, the
MLP module consists of a fully connected ReLU layer with $4096$ units terminating in a sigmoid layer giving the single probability score as the final model output.

All models are implemented in PyTorch and uses Adam optimisation~\cite{kingma2015} with a learning rate of $1\cdot{10}^{-4}$.

\subsection{Evaluation}
\label{subsect:evaluation}
We evaluate localisation performance for a query keyword $w$ as follows.
We first check whether the detection score ($\hat{p}_w$ or $\hat{y}_w$, depending on the model in Section~\ref{sect:models})
 of $w$ is greater or equal to a threshold $\theta$, and then select the position of the highest attention weight ($\alpha_t$ or $\alpha_{t, w}$) 
as the predicted location $\tau_p$ of $w$. $\tau_p$ is accepted as the correct location of a word if it falls within the interval corresponding to the true location of the word, according to forced alignments. The model is penalised whenever it localises a word whose detection score is less than $\theta$, i.e., if a word is not detected, then it is counted as a failed localisation even if the attention weight is at a maximum within that word. {We set the value of $\theta$ to $0.4$ and $0.5$ for BoW trained models and visually supervised models, respectively, which gave
best performance
on 
development data. 
We compute precision, recall, and $F1$ scores using $\tau_p$.\footnote{In \cite{palaz2016, olaleye2020}, this is referred to as the \textit{actual} metric.}

\section{Experimental results}

Table~\ref{tbl:localisation_evaluation} shows the keyword localisation performance 
for the attention-based models (CNN-Attend and CNN-PoolAttend), and the baseline models (PSC and CNN-Pool) when supervised with BoW labels and visual targets. The localisation scores are computed with detection scores taken into account (Section~\ref{subsect:evaluation}).
The goal is to see how well attention-based models perform on the keyword 
localisation task relative to 
models without attention. 
CNN-Attend trained with visual supervision
achieves a localisation precision 
of $31.2$\% and an $F1$ of $20.3\%$, 
outperforming the visually supervised PSC model.
However, we observe a contrasting results on the BoW-supervised models, where performance actually gets worse when comparing the PSC and CNN-Attend models. Comparing the BoW-supervised CNN-PoolAttend and CNN-Pool, we do see that attention gives better localisation than the GradCAM saliency-based localisation, but both of these models are outperformed by the PSC architecture.

The one reason that we did consider the convolutional architecture in CNN-PoolAttend and CNN-Pool (Section~\ref{subsec:model_variants})
is that it gives better 
scores than the other structure when only considering keyword detection, the task of predicting whether a keyword is present or not, irrespective of the localisation. This is shown in Table~\ref{tbl:detection_evaluation}.
But here we see that the visual CNN-Attend achieves close to the detection $F1$ of CNN-Pool ($28.8\%$ vs $29.2\%$) with the benefit of dramatically better localisation ($20.3\%$ vs $8.6\%$).
Although absolute performance is still modest, the performance of the
visually trained CNN-Attend suggests that attention-based CNN models is a better
 model structure for 
keyword localisation 
when hard ground truth labels
are not available.

To better understand the failure modes of the best visually supervised model, CNN-Attend, Figure~\ref{fig:qualitative_eval} shows localisation examples on test data. 
Figure~\ref{fig:qualitative_eval}(a) is an example of a case where the visually trained model successfully localises `white' although it is trained without explicit location information and hard ground truth text targets.
Figures~\ref{fig:qualitative_eval}(b), (c) and (d) are examples of failure cases. Figure~\ref{fig:qualitative_eval}(b) shows 
an outright failure. 
Figure~\ref{fig:qualitative_eval}(c) shows that when the model is prompted with `water', the attention weight is high on
`rowing', `rowboat' and `water', with `rowing' assigned the highest localisation score.
All of these words are semantically related to the query word.
Figure~\ref{fig:qualitative_eval}(d) shows that the model also sometimes localises
a single semantically related word in the utterance: the keyword `backstroke' is assigned highest localisation score when prompted with `swimming'.

To roughly quantify the different types of errors (of the types depicted in Figure~\ref{fig:qualitative_eval}), we randomly sample $100$ test utterances and visualise where the model attempts to locate a randomly selected keyword among the $67$ keywords in the system vocabulary.
We then categorise the $100$ visualisations into four groups. We find that, in this small sample set,} the model makes $48\%$ correct localisations, $39\%$ outright incorrect localisations, $7\%$ single-word semantic localisations, and $6\%$ multi-word semantic localisations, where the attention fires on the target word but also on multiple other words as in Figure~\ref{fig:qualitative_eval}(d).
The outright mistakes are therefore still high, but roughly $13\%$ of the mistakes can be seen as semantically related.

\section{Conclusions}

We investigated whether keyword localisation in speech is possible with attention-based convolutional neural networks (CNNs) trained with visual context.
We specifically compared two CNN models with attention to two non-attention baseline models used in previous work.
We found that adding attention 
improved localisation performance, but scores are still far from models trained with an unordered bag-of-word supervision (using transcriptions).
We qualitatively showed that the visually trained models sometimes locate semantically related words. 
Future work can examine cases where the models are evaluated not just on their ability to propose location of the exact keyword of interest, but also 
words that are semantically related 
to the keyword of interest.
Apart from our own preliminary work~\cite{olaleye2020}, the only other work that considered localisation with a visually supervised model is~\cite{harwath2017b, harwath2018a, william2019}. {But their 
 localisation at test time is based on having an image containing the word of interest (a text keyword is not provided)}. Future work could also see if insights from these two settings can be combined.

\vspace{2pt}
{\eightpt
\noindent \textbf{Acknowledgements.} This work is supported in part by the National Research Foundation of South Africa (grant no.\ 120409), a Google Faculty Award for HK, and a Google Africa PhD Scholarship for KO.
We would like to thank Benjamin van Niekerk for helpful input.}


\bibliographystyle{IEEEtran}

\bibliography{mybib}

\end{document}